\documentclass[letterpaper, 10 pt, conference]{ieeeconf}  
\usepackage{amsmath}
\usepackage{amssymb}
\usepackage{booktabs}
\usepackage{caption}
\usepackage{comment}
\usepackage{graphicx}
\usepackage{multirow}
\usepackage{pifont}
\usepackage[table]{xcolor}
\definecolor{darkgreen}{RGB}{0, 150, 0}
\definecolor{darkred}{RGB}{200, 0, 0}

\makeatletter
\let\NAT@parse\undefined
\makeatother

\newcommand{\iPi}{\Pi^{-1}}
\newcommand{\etal}{\textit{et al.}}

\newcommand{\ie}{\emph{i.e.}}
\newcommand{\ch}{{\color{darkgreen} \ding{51}}}
\newcommand{\xm}{{\color{darkred} \ding{55}}}
\usepackage[colorlinks,pagebackref=true,citecolor=green,bookmarks=false,hypertexnames=true]{hyperref} 

\title{\LARGE \bf
UnRectDepthNet: Self-Supervised Monocular Depth Estimation using a Generic Framework for Handling Common Camera Distortion Models
}

\author{
Varun Ravi Kumar$^{1,4\dag}$, 
Senthil Yogamani$^{2\dag}$, 
Markus Bach$^{1}$,\\
Christian Witt$^{1}$, 
Stefan Milz$^{3,4}$ and 
Patrick M\"ader$^{4}$\\
$^{1}$Valeo DAR Kronach, Germany \hspace{0.2cm}
$^{2}$Valeo Vision Systems, Ireland \\
$^{3}$Spleenlab GmbH, Germany \hspace{0.2cm}
$^{4}$Technische Universit\"at Ilmenau, Germany \hspace{0.2cm}
$^\dag$co-first authors
}

\begin{document}
\maketitle
\thispagestyle{empty}
\pagestyle{empty}
\begin{abstract}
In classical computer vision, rectification is an integral part of multi-view depth estimation. It typically includes epipolar rectification and lens distortion correction. This process simplifies the depth estimation significantly, and thus it has been adopted in CNN approaches. However, rectification has several side effects, including a reduced field of view (FOV), resampling distortion, and sensitivity to calibration errors. The effects are particularly pronounced in case of significant distortion (e.g., wide-angle fisheye cameras). In this paper, we propose a generic scale-aware self-supervised pipeline for estimating depth, euclidean distance, and visual odometry from unrectified monocular videos. We demonstrate a similar level of precision on the unrectified KITTI dataset with barrel distortion comparable to the rectified KITTI dataset. The intuition being that the rectification step can be implicitly absorbed within the CNN model, which learns the distortion model without increasing complexity. Our approach does not suffer from a reduced field of view and avoids computational costs for rectification at inference time. To further illustrate the general applicability of the proposed framework, we apply it to wide-angle fisheye cameras with 190$^\circ$ horizontal field of view. The training framework \emph{UnRectDepthNet} takes in the camera distortion model as an argument and adapts projection and unprojection functions accordingly. The proposed algorithm is evaluated further on the KITTI rectified dataset, and we achieve state-of-the-art results that improve upon our previous work FisheyeDistanceNet~\cite{kumar2019fisheyedistancenet}. Qualitative results on a distorted test scene video sequence indicate excellent performance\footnote{\url{https://youtu.be/K6pbx3bU4Ss}}.
\end{abstract}
 
\section{\textbf{Introduction}}

Depth estimation is a crucial task for automated driving, and multi-view geometric approaches were traditionally used for computing depth. Some of the initial prototypes of automated driving relied primarily on depth estimation~\cite{franke1998autonomous}, and to enable accurate depth estimation, stereo cameras were used. CNN models have been dominant with supervised learning in various visual perception tasks. Self-supervised learning has enabled high accuracy in depth estimation~\cite{godard2019digging, zhou2017unsupervised, monodepth17, Wang_2018_CVPR, varun18}. There is also a trend of integrating depth estimation task into multi-task models~\cite{sistu2019neurall, auxnet_visapp19}. Most of the depth estimation methods were demonstrated in the context of automated driving on rectified KITTI video sequences where barrel distortion was removed.\par

Rectification is considered to be a fundamental step in dense depth estimation~\cite{hartley2003multiple}. In the case of stereo cameras, epipolar rectification is performed to enable matching only in one direction along the horizontal scanline. This approach can also be extended to monocular cameras by using two consecutive frames giving rise to motion stereo. These rectification steps also require the removal of non-linear distortion. Although it is convenient to work with rectilinear projections, there are practical issues that arise due to rectification, which are discussed in detail in Section \ref{sec:problems}. 
\begin{figure}[!t]
  \captionsetup{singlelinecheck=false, font=small, labelsep=space}
  \centering
  \newcommand{\turnwidth}{0.485\columnwidth}

\newcommand{\imlabel}[2]{\includegraphics[width=0.488\columnwidth]{#1}
\raisebox{2pt}{\makebox[-2pt][l]{\footnotesize \sffamily #2}}}

\begin{tabular}{@{\hskip 0mm}c@{\hskip 1.5mm}c}
\centering
    \imlabel{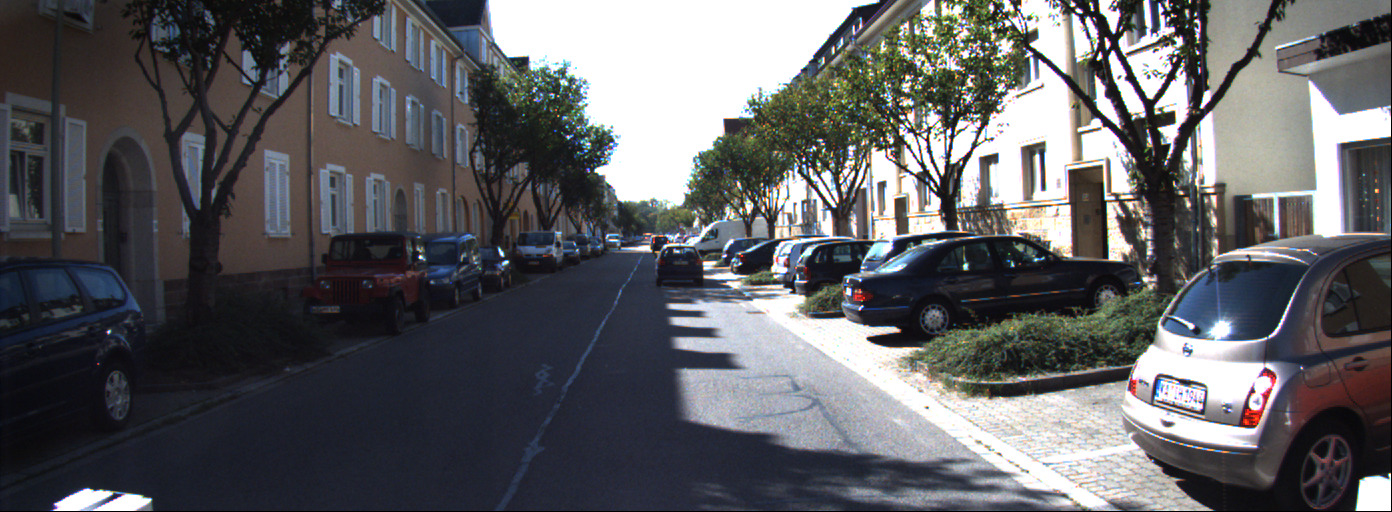}
    {\hspace{-0.47\columnwidth}\textcolor{white}{Unrectified}} &
    \imlabel{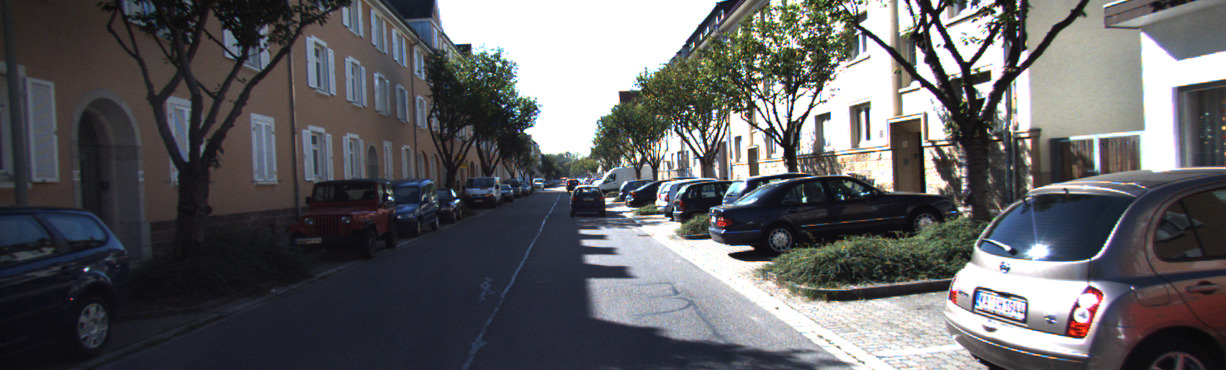}
    {\hspace{-0.47\columnwidth}\textcolor{white}{Rectified}} \\
    
    \imlabel{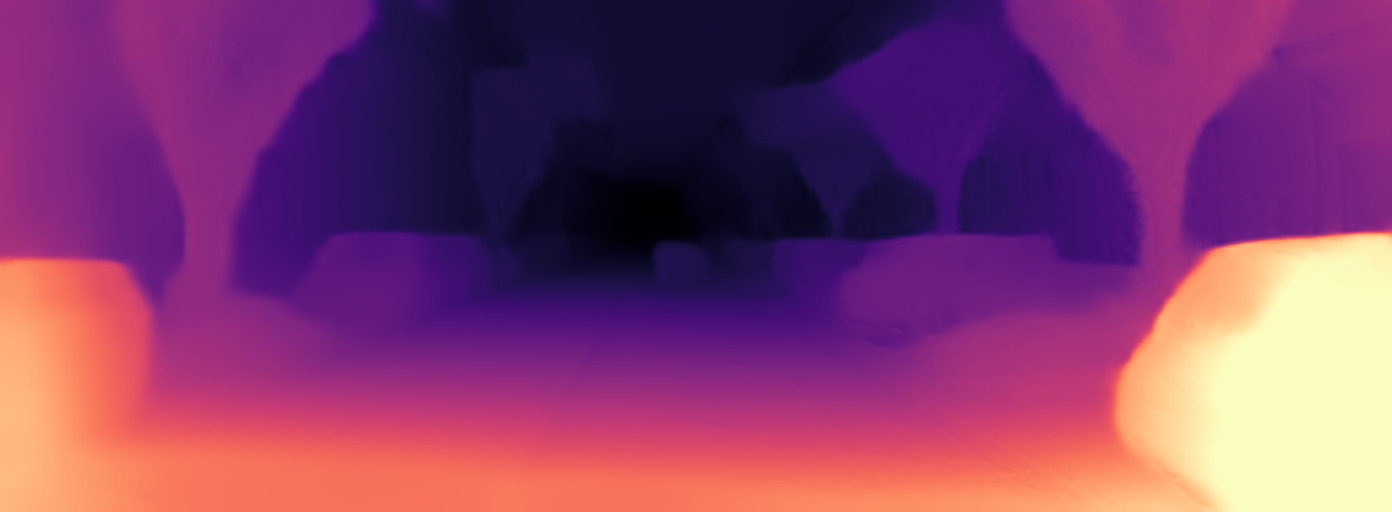}
    {} &
    \imlabel{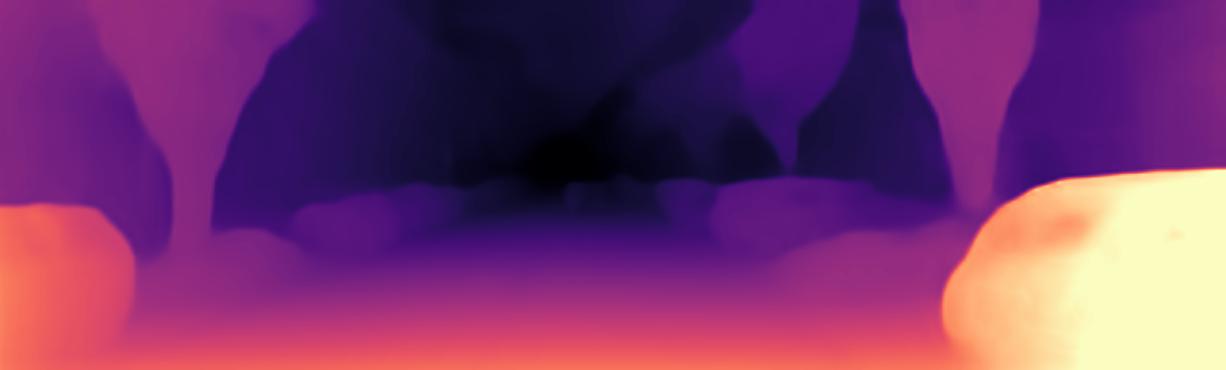}
    {} \\
\end{tabular}

  \caption{{\bf Depth obtained from a single unrectified (left) and  rectified KITTI image (right).} Our scale-aware model, UnRectDepthNet, yields precise boundaries and fine-grained depth maps.}
  \vspace{-1.8em}
  \label{fig:overview}
\end{figure}
Rectification has also been transferred to CNN based approaches as an inductive bias to simplify the learning. Yadati \etal~\cite{yadati2017multiscale} demonstrate that CNN based two-view depth estimation is challenging without rectification and attempt to solve it in a simpler setting. To the best of our knowledge, all the methods reported on KITTI make use of the barrel distortion corrected images. Automotive cameras like fisheye surround-view cameras exhibit a strong distortion, and it is not easy to rectify their images. Recently, several tasks such as motion segmentation~\cite{yahiaoui2019fisheyemodnet}, and soiling detection~\cite{uvrivcavr2019soilingnet} were demonstrated on fisheye images without rectification.

\begin{figure*}[t]
  \captionsetup{belowskip=0pt, font= small, singlelinecheck=false}
    \centering
    \includegraphics[width=\textwidth]{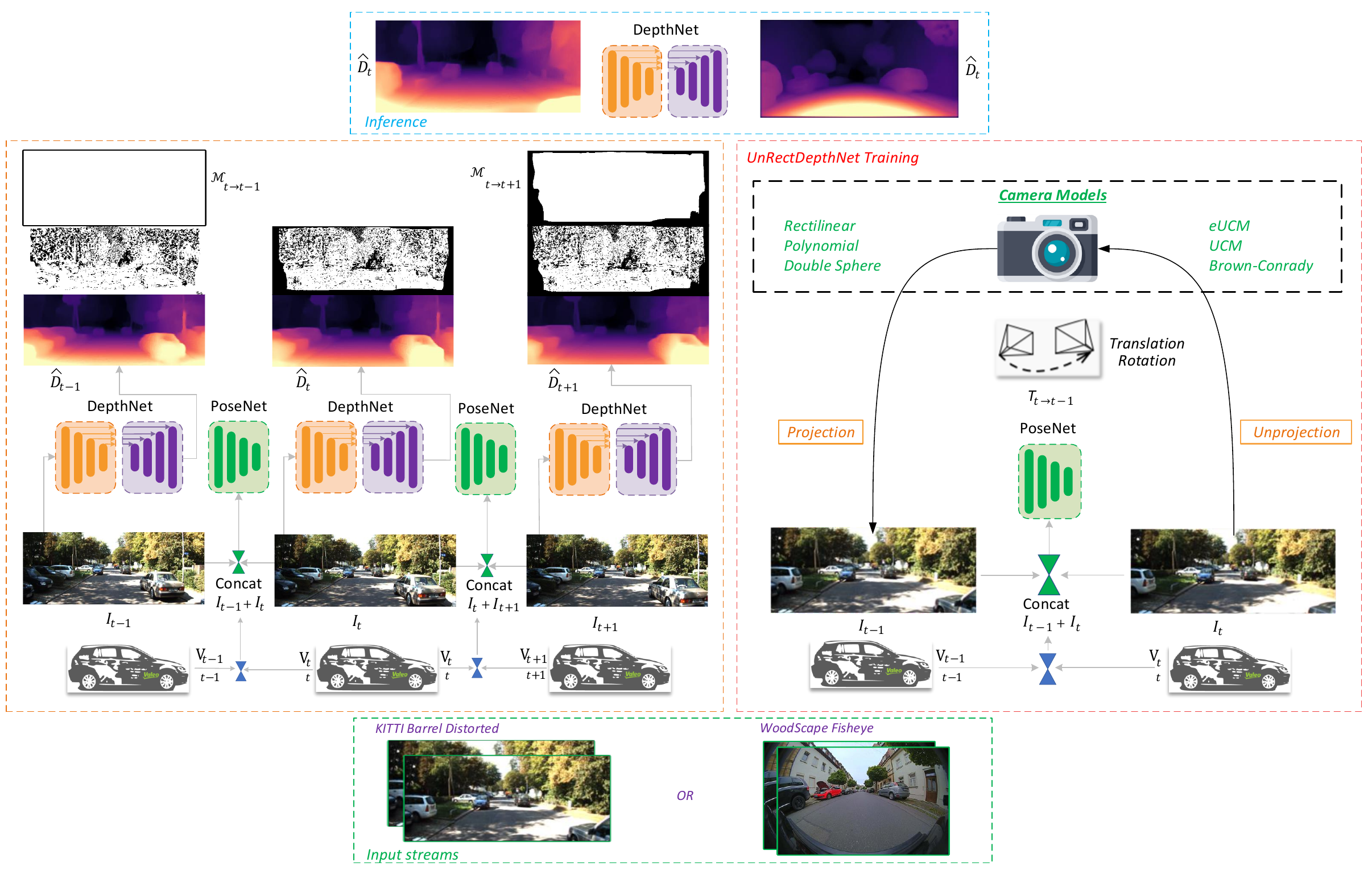}
    \caption{\textbf{Overview of our generic depth estimation training framework  UnRectDepthNet.}
     The UnRectDepthNet training block on the right enables the usage of various camera models generically listed in the black box. The distortion is then handled internally in the unprojection and projection steps of the transformation from $I_t$ to $I_{t-1}$. In this paper, we have tested it with KITTI barrel distorted and WoodScape fisheye distorted video sequences. The block on the left indicates the entire workflow of the training pipeline where the top row depicts the ego masks as explained in~\cite{kumar2019fisheyedistancenet}, $\mathcal{M}_{t \to t-1}$, $\mathcal{M}_{t \to t+1}$ representing the valid pixel coordinates while synthesizing $\hat I_{t-1 \to t}$ from $I_{t-1}$ and $\hat I_{t+1 \to t}$ from $I_{t+1}$ respectively. The following row showcases the masks used to filter static pixels, obtained after training two epochs, and the black pixels are removed from the reconstruction loss. Dynamic objects moving at speed similar to the ego car's as well as homogeneous areas are filtered out to prevent the contamination of reconstruction loss. The third row shows the depth predictions, where the scale ambiguity is resolved using the ego vehicle's odometry data. Finally, the top block illustrates the inference output.
     }
    \label{fig:model_arch}
    \vspace{-0.4cm}
\end{figure*}

In our previous article, FisheyeDistanceNet~\cite{kumar2019fisheyedistancenet}, we introduced the first end-to-end scale-aware, self-supervised monocular training method for fisheye cameras with a large field of view to regress a Euclidean distance map. In this work, we generalize the training framework to work with any camera model and propose a fully-differentiable architecture that estimates the depth directly from raw unrectified images (shown in Fig.~\ref{fig:model_arch}) without the need for any pre-processing. In terms of the motivation of a generic training pipeline, the closest related work is CAM-Convs~\cite{facil2019cam}, where authors propose a generic framework for different types of cameras. However, they do not handle non-linear distortion, and it is not a self-supervised method. Our contributions include:
\begin{itemize}
\item A novel generic end-to-end self-supervised training pipeline to estimate monocular depth maps on raw distorted images for various camera models.
\item Empirical evaluation of our approach on two diverse automotive datasets, namely KITTI and WoodScape.
\item First demonstration of depth estimation results directly on unrectified KITTI sequences (see Fig.~\ref{fig:overview}).
\item State-of-the-art results on KITTI depth estimation among self-supervised methods.
\end{itemize}
\section{\textbf{Motivation for Working on Raw Images}}

\textit{\textbf{Distortion in Automotive Cameras}}
\label{sec:distortion_models}
To handle the wide variety of automotive use cases, different cameras having a different field of view are used. The most common ones are around $100^\circ$ hFOV (horizontal field of view) cameras used for front camera sensing and $190^\circ$ hFOV fisheye lens cameras for surround-view sensing. Due to their moderate to large FOV, these cameras suffer from lens distortion, whose main component is typically radial distortion and minor tangential distortion.\par
\textit{\textbf{Moderate FOV Lens Model}}
\label{sec:large_fov}
For lenses with a moderate FOV (${<120^\circ}$), Brown–Conrady model \cite{conrady1919decentred} is commonly used as it models both radial and tangential distortion. For larger FOV, this distortion model typically breaks down or requires very high polynomial orders. The KITTI dataset's calibration uses this model based on OpenCV's implementation. In this model, the projection function $X_c \mapsto \Pi(X_c) = p$ maps a 3D point $X_c = (x_c, y_c, z_c)^T$ in camera coordinates to a pixel $p = (i, j)^T$ in the image coordinates. It is calculated in the following way:
\begin{gather*}
   x = x_c / z_c,\quad y = y_c / z_c \\
   x' = x (1 + k_1 r^2 + k_2 r^4 + k_3 r^6) + 2 p_1 x y + p_2(r^2 + 2 x^2) \\
   y' =  y (1 + k_1 r^2 + k_2 r^4 + k_3 r^6) + p_1 (r^2 + 2 y^2) + 2 p_2 x y \\
   i = f_x \cdot x' + c_x,\quad
   j = f_y \cdot y' + c_y 
\end{gather*}
where $k_1$, $k_2$, and $k_3$ are radial distortion coefficients, $p_1$ and $p_2$ are tangential distortion coefficients of the lens, $r^2 = x^2 + y^2$, $f_x$, $f_y$ are the focal lengths and $c_x$, $c_y$ are the coordinates of the principal point.\par
\textit{\textbf{Fisheye Lens Models:}} For fisheye lenses, the mapping of 3D points to pixels universally requires a radial component $r\left(\theta\right)$ \cite{hughes2010fisheye}. The projection is a complex multi-stage process compared to regular lenses and thus we list the detailed steps:
\begin{enumerate} 
    \item The point $X_c$ in camera coordinates is mapped to a unit vector
    as $S = (s_x, s_y, s_z)^T = X_c / \|X_c\|$.
    \item The incident angle against the optical axis (coincident with the $Z$-axis) $\theta = \frac{\pi}{2} - \arcsin\left(s_z\right)$ is computed.
    \item The radial function $r(\theta)$ to get the radius on the image plane (typically in pixels) is computed.
    \item Given the pixel distortion centre $(c_x,c_y)$, the pixel location is given by $i = r\cdot s_x / \rho + c_x$ and $j = r\cdot s_y / \rho + c_y$ with $\rho = \sqrt{s_x^2 + s_y^2}$. 
    \item (optional) Depending on the model used in Step 3, an additional distortion correction may need to be applied.
\end{enumerate} 

We discuss the projection models which are supported in our framework. The polynomial model is the commonly used one and relatively recent projection models are UCM (Unified Camera Model) \cite{barreto2006unified} and eUCM (Enhanced UCM) \cite{Khomutenko2016eucm}. Rectilinear (representation of pinhole model) and Stereographic (mapping of sphere to a plane) are not suitable for fisheye lenses but provided for comparison. Double Sphere \cite{usenko2018double} is a recently proposed model which has a closed form inverse with low computational complexity. The radial distortion models are summarized below:
\begin{enumerate}
\item Polynomial: $r(\theta) = a_1 \theta + a_2 \theta^2 + a_3 \theta^3 + a_4 \theta^4$
\item UCM: $r(\theta) =  f\cdot\sin\theta / (\cos\theta + \xi)$
\item eUCM: $r(\theta) = f\cdot\frac{\sin\theta}{\cos\theta + \alpha\left(\sqrt{\beta\cdot \sin^2\theta + \cos^2\theta} - \cos\theta\right)}$ 
\item Rectilinear: $r(\theta) = f \cdot \tan\theta$
\item Stereographic: $r(\theta) = 2 f \cdot \tan(\theta/2)$ 
\item Double Sphere:\newline $r(\theta) = f\cdot \frac{\sin\theta}{\alpha\sqrt{\sin^2\theta + (\xi + \cos\theta)^2} + (1-\alpha)(\xi + \cos\theta)}$
\end{enumerate} 
\begin{figure}[!t]
\vspace{0.15cm}
  \captionsetup{singlelinecheck=false, font=small, labelsep=space, belowskip=-8pt}
  \centering
  \newcommand{\turnwidth}{0.485\columnwidth}

\newcommand{\imlabel}[2]{\includegraphics[width=0.488\columnwidth]{#1}
\raisebox{2pt}{\makebox[-2pt][r]{\footnotesize #2}}}

\begin{tabular}{@{\hskip 0mm}c@{\hskip 1.5mm}c}
\centering
    \imlabel{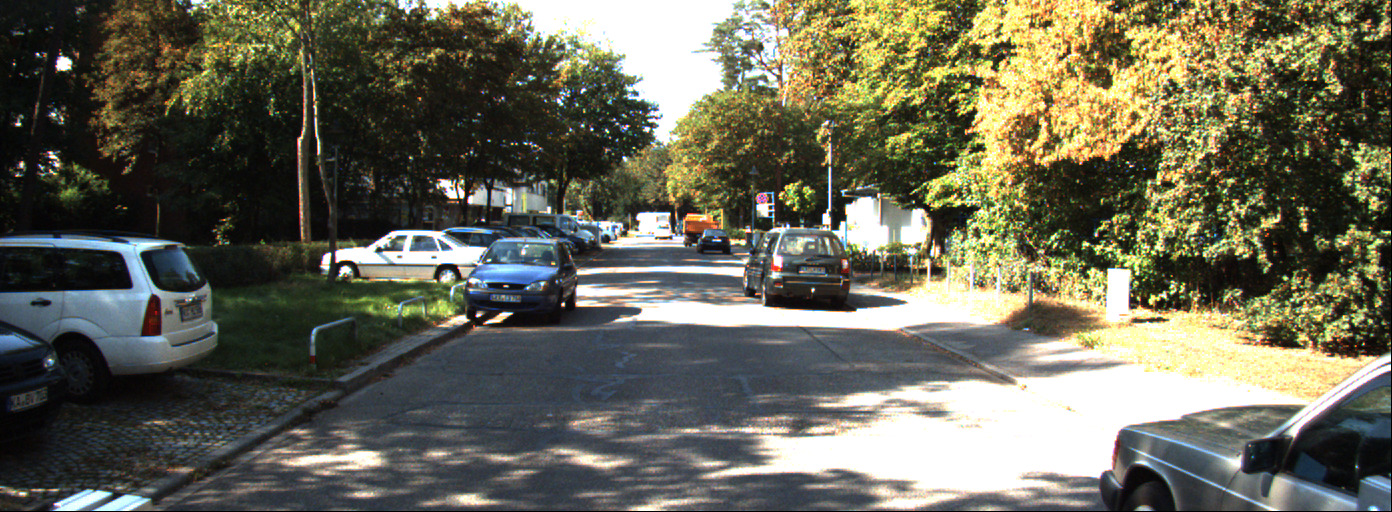} &
    \imlabel{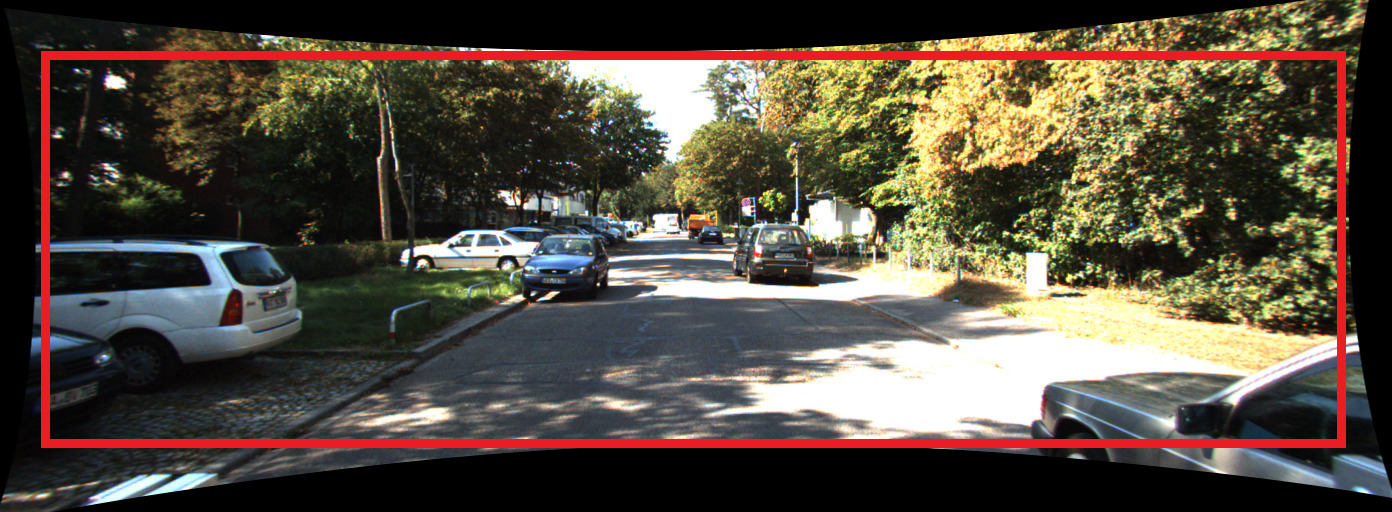}
    {} \\
    \imlabel{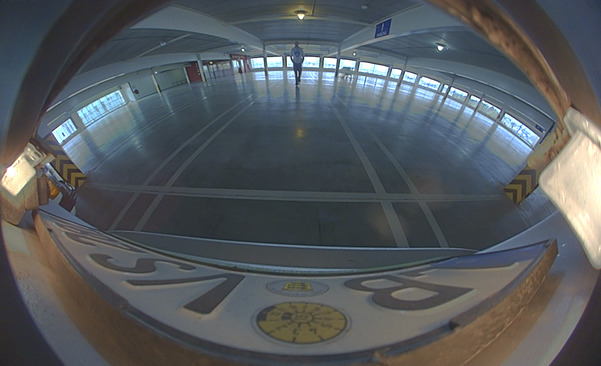}
    {} &
    \imlabel{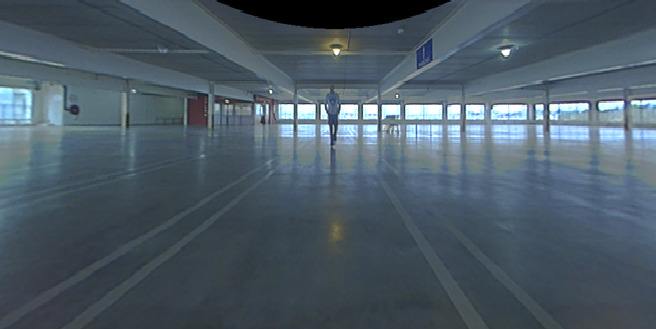}
    {} \\
\end{tabular}
  \caption{\textbf{Illustration of distortion correction in KITTI and WoodScape datasets.} The first row shows a raw KITTI image with barrel distortion and the corresponding rectified image. The red box was used to crop out black pixels in periphery causing a loss of FOV. The second row shows a raw WoodScape image with strong fisheye lens distortion and the corresponding rectified image exhibiting a drastic loss of FOV.}
  \label{fig:undistort}
\end{figure}
\textit{\textbf{Practical Problems encountered}}
\label{sec:problems}
In the previous subsection, we have established that real-world automotive cameras have lens distortion. The typical approach is to remove the distortion and then apply standard camera projection models. However, in practice, this has several issues that are not dealt with in the literature. Fig.~\ref{fig:undistort} illustrates the rectification used in KITTI and WoodScape datasets.  In the KITTI dataset, due to the barrel\footnote{KITTI \cite{geiger2013vision} refers to it as pincushion distortion because of an error in the OpenCV documentation which was fixed later.}  distortion effect of the camera, images have been rectified and cropped to $1242\times375$ pixels. Cropping is performed after rectification to get a rectangular grid without any black pixels in the periphery. Thus the size of the rectified images is smaller than that of the raw images with $1392\times512$ pixels.
Based on the number of the non-black, occupied pixels removed by the cropping, roughly $10\%$ of the image information is lost. This effect becomes more drastic for the WoodScape images with a much larger radial distortion were more than $30\%$ of the image information is lost\footnote{Other quasi-linear rectification methods like cylindrical rectification will preserve more information at the cost of additional distortion.}. For a horizontal FOV greater than $180^\circ$, there are rays incident from behind the camera, making it theoretically impossible to establish a complete mapping to a rectilinear viewport. Thus the rectification defeats the purpose of using a wide-angle fisheye lens.

Reduced FOV is the most critical problem of undistortion, but there are further practical issues. The first one is resampling distortion, which is caused by interpolation errors during the warping step. This effect can be partially mitigated by a more advanced interpolation method \cite{meijering2002chronology}. However, it is particularly strong in the periphery of fisheye lenses because a small region is expanded to a larger one in the warped image. Besides, the warping step is needed at inference time, which consumes significant computing power and memory bandwidth. 

The other issue is related to calibration. In an industrial setup, millions of cameras are deployed, and they have manufacturing variations. The camera parameters (mainly focal length) can also vary due to high ambient temperatures when driving in a hot region. Thus a model that relies on rectification to correct the distortion could have errors. For instance, dataset capture and training are typically performed on one particular camera, and the model is deployed to work on millions of cameras in commercial vehicles. Thus rectification and cropping to a standard resolution as per the training camera are sub-optimal for a deployed camera. However, if CNN learns the distortion as part of the transfer function, it is only weakly encoded and thus expected to be more robust. To alleviate these issues, we are motivated to explore a depth estimation model that can work directly on raw images without needing rectification.
\section{\textbf{Self-Supervised Scale-Aware Depth Estimation}}

Following Zhou \etal~\cite{zhou2017unsupervised} we aim at learning a self-supervised monocular structure-from-motion (SfM):
\begin{enumerate}
\item a scale-ambiguous depth $\hat D$ is obtained through a self-supervised monocular model $g_D: I_t \to D$ outputting $\hat D = g_D(p)$ per pixel $p$ in the target image $I_t$; and
\item Rigid transformations $T_{t \rightarrow t'} \in \text{SE(3)}$ which comprise a set of 6 degrees of freedom are estimated using an ego-motion predictor $g_x: (I_t, I_{t^\prime}) \to I_{t \to t^\prime}$, between the target image $I_t$ and the set of reference images $I_{t^\prime}$. Specifically, $t' \in \{t+1, t-1\}$, \ie the frames $I_{t-1}$ and $I_{t+1}$ are used as reference images, but it would in general be possible to use a larger offset from the temporal consistent sequence.
\end{enumerate}

In all the previous works~\cite{zhou2017unsupervised, godard2019digging, guizilini2019packnet}, networks are equipped to retrieve inverse depth $g_d: p \mapsto 1/g_D(p)$. One downside to these methods is the scale ambiguity in both depth and pose estimation. In this work, we recover scale-aware depth directly for distorted images. View-synthesis is used as a self-supervising technique, and the network is trained with the source images $I_{t-1}$ and $I_{t+1}$ to synthesize the appearance of a target image $I_t$. For this, we need the projection function $\Pi$ of the chosen camera model, which maps a 3D point $X_c$ in camera coordinates to a pixel $p = \Pi(X_c)$ in image coordinates. An overview of projection models for different lens types can be found in Section~\ref{sec:distortion_models}. The corresponding unprojection function $\iPi$, which maps an image pixel $p$ and its depth estimate $\hat D$ to the 3D point $X_c = \iPi(p,\hat D)$, is also required. If $\iPi$ cannot be expressed in analytic form, a pre-calculated lookup table is used to ensure computational efficiency.\par
\textit{\textbf{View Synthesis for Various Camera Models}} \label{sec:view_synthesis}
Using the network's $\hat D_t$ depth estimate for frame $I_t$ at time $t$ a point cloud $P_t$ is obtained through:
\begin{equation} \label{eq:pt}
    P_t = \iPi(p_t, \hat D_t)
\end{equation}
Here, the pixel set of image $I_t$ is represented by $p_t$ and $\iPi$ denotes the unprojection function introduced above. The pose of target image $I_t$ relative to the pose of the source image $I_{t^\prime}$ is considered as $T_{t\to t^\prime}$ and estimated by the pose network. For the point cloud of frame $I_{t^\prime}$ an estimate $\hat{P}_{t'} = T_{t \to t^\prime} P_t$ is obtained by applying this transformation. The projection model $\Pi$ is used at time $t'$ to project $\hat{P}_{t^\prime}$ onto the camera. The projection and transformation are combined with Eq.~\ref{eq:pt} to establish a mapping between the image coordinates $p_t=(i,j)^T$ at time $t$ and $\hat p_{t^\prime}=(\hat i, \hat j)^T$ at time $t^\prime$. A view-synthesized reconstruction $\hat I_{t' \to t}$ of the target frame $I_t$ is computed via a backward warp of the source frame $I_{t^\prime}$ using this mapping.
\begin{align}
    \label{pixel_estimate}
    \hat{p}_{t'} = \Pi \big( T_{t \to t'} \iPi(p_t,\hat D_t) \big), \quad
    \hat{I}_{t' \to t}^{ij} = \Big\langle I_{t'}^{\hat{i}\hat{j}} \Big\rangle
\end{align}
Because the warped coordinates $\hat i, \hat j$ are continuous,
differentiable spatial transformer network~\cite{jaderberg2015spatial} can be used to synthesize $\hat{I}_{t'\to t}$ by bilinear interpolation of the four pixels of $I_{t^\prime}$ nearby $\hat p_{t'}$. The $\big\langle\dots\big\rangle$ symbol denotes the corresponding operator used for sampling.\par
\textit{\textbf{Reconstruction Loss}} \label{sec:reconstruction loss}
The L1 pixel-wise loss is coupled with Structural Similarity (SSIM)~\cite{wang2004image}, to form an image reconstruction loss $\mathcal{L}_{r}$ between the target image $I_t$ and the reconstructed target image $\hat I_{t' \to t}$ given by Eq.~\ref{eq:loss-photo} below.
\begin{align}
  \label{eq:loss-photo}
  \tilde{\mathcal{L}}_{r}(I_t,\hat I_{t' \to t}) &= \omega~\frac{1 - \text{SSIM}(I_t,\hat I_{t' \to t}, \mathcal{M}_{t \to t^\prime})}{2} \nonumber \\
  &\quad+ (1-\omega)~ \left\| (I_t - \hat I_{t' \to t}) \odot \mathcal{M}_{t \to t^\prime} \right\|_{l^1} \nonumber \\
    \mathcal{L}_{r} &= \min_{t^\prime \in \{t+1,t-1\}} \tilde{\mathcal{L}}_r(I_t, \hat I_{t' \to t})
\end{align}
The binary mask $\mathcal{M}_{t \to t^\prime}$ from~\cite{kumar2019fisheyedistancenet} is incorporated, element-wise multiplication is denoted by $\odot$ and $\omega$ is set to 0.85. Following~\cite{godard2019digging}, we adopt a per-pixel minimum compared to averaging over all the source images. This yields higher accuracy by reducing the artifacts and significantly sharpening the boundaries of occlusion. We clip the reconstruction loss values to a $95^\text{th}$ percentile based on~\cite{zhou2018unsupervised} to diminish the impact of dynamic objects or occluded areas in the scene. It indirectly aids the optimization algorithm to have a robust reconstruction error.\par
\textit{\textbf{Solving Scale Factor Ambiguity at Training Time}} \label{sec: scale-aware sfm}
The network's sigmoid output $\sigma$ can be translated to depth with $D = 1 / ({x\sigma + y})$ based on the fact that $depth \propto 1 / disparity$ for a rectified pinhole projection model, where $x$ and $y$ are chosen to constrain $D$ between $0.1$ and $100$ units~\cite{godard2019digging}. We can only obtain angular disparities~\cite{arican2009dense} for camera models which undergo distortion. To perform a successful inverse warp operation of source images $I_{t'}$ onto the target frame $I_t$, metric depth values are required. \textit{Scale-ambiguous} estimates from both the $g_d$ monocular depth model and the $g_\mathbf{x}$ ego-motion predictor is a hindrance to the obtainment of metric depth maps on any camera model of choice because of the inherent drawback of the self-supervised structure-from-motion objective. Following~\cite{kumar2019fisheyedistancenet}, we solve the scale ambiguity by normalizing the pose network's prediction $T_{t \to t'}$ to obtain scale-aware depth values. We compute the displacement magnitude $\Delta x$ relative to target frame $I_t$ utilizing the ego vehicle's instantaneous velocity predictions $v_{t'}$ at time $t'$ and $v_t$ at time $t$ obtained from its odometry data. Finally, we scale the normalized translation vector with $\Delta x$. The same method is also incorporated on KITTI~\cite{KITTIDepth} rectified pinhole dataset to achieve scale-aware depth maps.
\begin{align}
    \overline{T}_{t \to t'} = \frac {T_{t \to t'}} {\|T_{t \to t'}\|} \cdot \Delta x
\end{align}
\textit{\textbf{Edge-Aware Depth Smoothness Loss}}
A geometric smoothness loss is added to regularize depth and avoid different values in occluded or homogeneous areas. We incorporate the edge-aware loss term and impose it on the inverse depth map similar to~\cite{monodepth17, mahjourian2018unsupervised, zou2018df}.
\begin{align}
    \mathcal{L}_{s}(\hat{D}_t) = | \partial_i \hat{D}^*_t | e^{-|\partial_i I_t|} + | \partial_j \hat{D}^*_t | e^{-|\partial_j I_t|}
\end{align}
Following~\cite{Wang_2018_CVPR}, mean-normalized inverse depth $\hat{D}^*_t$ of the target image $I_t$ is considered to avoid any shrinkage of depth estimates $\hat D_t$, \ie $\hat{D}^*_t = \hat{D}^{-1}_t / \overline{D}_t$, where $\overline{D}_t$ denotes the mean of $\hat{D}^{-1}_t := 1 /\hat{D}_t$.\par
\begin{table*}[!ht]
\captionsetup{belowskip=-8pt, font= small, singlelinecheck=false}
\renewcommand{\arraystretch}{0.87}
\centering
{
\small
\setlength{\tabcolsep}{0.3em}
\begin{tabular}{c|lccccccccc}
\toprule
& \textbf{Method} & Resolution & Dataset & \cellcolor[HTML]{5880ab}Abs Rel & \cellcolor[HTML]{5880ab}Sq Rel & \cellcolor[HTML]{5880ab}RMSE & \cellcolor[HTML]{5880ab}RMSE$_{log}$ & \cellcolor[HTML]{e8715b}$\delta<1.25$ & \cellcolor[HTML]{e8715b}$\delta<1.25^2$ & \cellcolor[HTML]{e8715b}$\delta<1.25^3$ \\
\cmidrule(lr){5-8} \cmidrule(lr){9-11}
& & & & \multicolumn{4}{c}{\cellcolor[HTML]{5880ab}lower is better} & \multicolumn{3}{c}{\cellcolor[HTML]{e8715b}higher is better}\\
\toprule
\parbox[t]{2mm}{\multirow{14}{*}{\rotatebox[origin=c]{90}{Original~\cite{Eigen_14}}}}
& SfMLeaner~\cite{zhou2017unsupervised}            & 416 x 128  & K  & 0.183 & 1.595 & 6.709 & 0.270 & 0.734 & 0.902 & 0.959 \\
& Vid2depth~\cite{mahjourian2018unsupervised}      & 416 x 128  & K  & 0.163 & 1.240 & 6.220 & 0.250 & 0.762 & 0.916 & 0.968 \\
& DDVO~\cite{Wang_2018_CVPR}                       & 416 x 128  & K  & 0.151 & 1.257 & 5.583 & 0.228 & 0.810 & 0.936 & 0.974 \\
& EPC++~\cite{luo2018every}                        & 640 x 192  & K  & 0.141 & 1.029 & 5.350 & 0.216 & 0.816 & 0.941 & 0.976 \\
& Struct2Depth~\cite{casser2019depth}              & 416 x 128  & K  & 0.141 & 1.026 & 5.291 & 0.215 & 0.816 & 0.945 & 0.979 \\
& Monodepth2~\cite{godard2019digging}              & 640 x 192  & K  & 0.115 & 0.903 & 4.863 & 0.193 & 0.877 & 0.959 & 0.981 \\ 
& PackNet-SfM~\cite{guizilini2019packnet}          & 640 x 192  & K  & 0.111 & 0.785 & 4.601 & 0.189 & 0.878 & 0.960 & 0.982 \\
& Monodepth2~\cite{godard2019digging}              & 1024 x 320 & K  & 0.115 & 0.882 & 4.701 & 0.190 & 0.879 & 0.961 & 0.982 \\
\cmidrule{2-11}
& \textbf{UnRectDepthNet}                          & 640 x 192  & K  & \textbf{0.107} & \textbf{0.721} & \textbf{4.564} &\textbf{0.178} & \textbf{0.894} & \textbf{0.971} & \textbf{0.986} \\
& \textbf{UnRectDepthNet}                          & 1024 x 320 & K  & 0.103 & 0.705 & 4.386 & 0.164 & 0.897 & 0.980 & 0.989 \\
& \textbf{UnRectDepthNet}                          & 608 x 224  & KD & 0.102 & 0.720 & 4.559 & 0.183 & 0.892 & 0.973 & 0.988 \\
& \textbf{UnRectDepthNet}                          & 1216 x 448 & KD & 0.106 & 0.709 & 4.357 & 0.161 & 0.895 & 0.984 & 0.992 \\
& FisheyeDistanceNet~\cite{kumar2019fisheyedistancenet} & 512 x 256  & WS & 0.152 & 0.768 & 2.723 & \textbf{0.210} & 0.812 & 0.954 & 0.974 \\
& \textbf{UnRectDepthNet}                          & 512 x 256  & WS & \textbf{0.148} & \textbf{0.702} & \textbf{2.530} & 0.212 & \textbf{0.826} & \textbf{0.960} & \textbf{0.980} \\
\midrule
\parbox[t]{2mm}{\multirow{8}{*}{\rotatebox[origin=c]{90}{Improved~\cite{uhrig2017sparsity}}}}
& SfMLeaner~\cite{zhou2017unsupervised}            & 416 x 128 & K & 0.176 & 1.532 & 6.129 & 0.244 & 0.758 & 0.921 & 0.971 \\
& Vid2Depth~\cite{mahjourian2018unsupervised}      & 416 x 128 & K & 0.134 & 0.983 & 5.501 & 0.203 & 0.827 & 0.944 & 0.981 \\
& DDVO~\cite{Wang_2018_CVPR}                       & 416 x 128 & K & 0.126 & 0.866 & 4.932 & 0.185 & 0.851 & 0.958 & 0.986 \\
& EPC++~\cite{luo2018every}                        & 640 x 192 & K & 0.120 & 0.789 & 4.755 & 0.177 & 0.856 & 0.961 & 0.987 \\
& Monodepth2~\cite{godard2019digging}              & 640 x 192 & K & 0.090 & 0.545 & 3.942 & 0.137 & 0.914 & 0.983 & 0.995 \\
& PackNet-SfM~\cite{guizilini2019packnet}          & 640 x 192 & K & \textbf{0.078} & 0.420 & 3.485 & 0.121 & \textbf{0.931} & 0.986 & \textbf{0.996} \\
\cmidrule{2-11} 
& \textbf{UnRectDepthNet}                          & 640 x 192 & K  & 0.081 & \textbf{0.414} & \textbf{3.412} & \textbf{0.117} & 0.926 & \textbf{0.987} & \textbf{0.996} \\
& \textbf{UnRectDepthNet}                          & 640 x 224 & KD & 0.092 & 0.458 & 3.503 & 0.132 & 0.906 & 0.971 & 0.990 \\
\bottomrule
\end{tabular}
}
\caption{\textbf{Quantitative performance comparison of UnRectDepthNet} for depths up to 80\,m for KITTI and 40\,m for WoodScape. In the Dataset column, K refers to KITTI~\cite{KITTIDepth}, KD refers to the KITTI distorted~\cite{geiger2013vision}, and WS refers to WoodScape~\cite{yogamani2019woodscape} dataset. \textit{Original} refers to  depth maps defined in \cite{Eigen_14}, and \textit{Improved} refers to refined depth maps provided by \cite{uhrig2017sparsity}. All the methods listed in the table are self-supervised approaches on monocular camera sequences. At inference time, all the approaches except UnRectDepthNet and PackNet-SfM scale the estimated depths using median ground-truth LiDAR depth. We generalized our previous model FisheyeDistanceNet in our new training framework and added additional features which improve results on WoodScape.}
\label{table:results}
\end{table*}
\begin{table*}[t!]
\captionsetup{skip=-2pt, belowskip=-10pt, font= small, singlelinecheck=false}
\small
\begin{center}
	\begin{tabular}{l|c|c|c|c|c|c|c|c|c|c|c}
	\toprule
    Method & FS & BS & SR & CSDCL & \cellcolor[HTML]{5880ab}Abs Rel & \cellcolor[HTML]{5880ab}Sq Rel & \cellcolor[HTML]{5880ab}RMSE & \cellcolor[HTML]{5880ab}RMSE$_{log}$ & \cellcolor[HTML]{e8715b}$\delta < 1.25$ & \cellcolor[HTML]{e8715b}$\delta < 1.25^2$ &  \cellcolor[HTML]{e8715b}$\delta < 1.25^3$\\
    \toprule
    Ours & \ch & \ch & \ch & \ch & 0.102 & 0.720 & 4.559 & 0.183 & 0.892 & 0.973 & 0.988 \\
    Ours & \ch & \xm & \ch & \ch & 0.131 & 0.856 & 4.933 & 0.198 & 0.853 & 0.954 & 0.968 \\
    Ours & \ch & \xm & \xm & \ch & 0.141 & 0.971 & 5.183 & 0.206 & 0.831 & 0.941 & 0.953 \\
    Ours & \ch & \xm & \xm & \xm & 0.144 & 1.011 & 5.204 & 0.225 & 0.822 & 0.945 & 0.949 \\
    \bottomrule
\end{tabular}
\end{center}
\caption{\textbf{Ablation study of our algorithm on the KITTI Eigen split dataset~\cite{eigen2015predicting}}. Depths are capped at 80\,m. FS, BS, SR, CSDCL indicate forward sequence, backward sequence, super-resolution network with PixelShuffle~\cite{aitken2017checkerboard} layers and cross-sequence depth consistency loss, respectively. The input resolution is $608\times224$ pixels.} 
\label{table:kittiablation}
\end{table*}
\textit{\textbf{Final Training Loss}}
The final self-supervised structure-from-motion (SfM) objective comprises a reconstruction loss $\mathcal{L}_r$ applied on forward and backward sequences and an edge-aware smoothness term $\mathcal{L}_s$ to regularize depth. Additionally, $\mathcal{L}_{dc}$ a cross-sequence depth consistency loss estimated from the sequence of frames in the training videos is also incorporated just as in~\cite{kumar2019fisheyedistancenet}. Since the bilinear sampler has gradient locality~\cite{jaderberg2015spatial}, we include four scales for training as suggested in~\cite{zhou2017unsupervised,monodepth17} mainly to reduce the chance of training reaching a local minimum. The overall objective function is averaged over the number of pixels, scales and batches. 
\begin{align} 
    \label{equ:objective}
    \mathcal{L} &= \sum\limits_{n = 1}^4 {\frac{{\mathcal{L}_{n}}}{{{2^{n - 1}}}}} ,\\
    \mathcal{L}_{n} &= {}^n\mathcal{L}_{r}^f + {}^n\mathcal{L}_{r}^b + \gamma\ {}^n{\mathcal{L}_{dc}} + \beta\ {}^n{\mathcal{L}_{s}} \nonumber
\end{align}
\section{\textbf{Implementation Details}}
The depth estimation network is mainly based on FisheyeDistanceNet~\cite{kumar2019fisheyedistancenet}. We use Pytorch~\cite{paszke2017automatic} and employ Ranger (RAdam \cite{liu2019variance} + LookAhead~\cite{zhang2019lookahead}) optimizer to minimize the training objective function (\ref{equ:objective}). The model is trained using Titan RTX with a batch size of 20 for 20 epochs, with initial learning rate of 4 x ${{10}^{-4}}$ with OneCycleScheduler \cite{smith2019super}. The network's sigmoid output $\sigma$ is converted to depth with $D = 1 / ({x \cdot \sigma + y})$ for pinhole model and $D = {x \cdot \sigma + y}$ for fisheye, where $x$ and $y$ are chosen such that $D$ is bounded between $0.1$ and $100$ units. For KITTI distorted images, we use $608\times224$ pixels, and for WoodScape fisheye images $512\times256$ pixels as the network input to maintain the original aspect ratio. The loss weighting factors $\beta$ and $\gamma$ of smoothness and cross-sequence depth consistency loss are set to $0.001$. To remove checkerboard artifacts in the sub-pixel convolution~\cite{shi2016real}, the final convolutional layers are initialized in a particular manner before the pixel shuffle operation as described in~\cite{aitken2017checkerboard}.
\begin{figure*}[!ht]
\captionsetup{belowskip=-12pt, font= small}
  \centering
  \resizebox{\textwidth}{!}{
  \newcommand{\turnheightnew}{0.25\columnwidth}
\centering

\begin{tabular}{@{\hskip 0.5mm}c@{\hskip 0.5mm}c@{\hskip 0.5mm}c@{\hskip 0.5mm}c@{\hskip 0.5mm}c@{}}

{\rotatebox{90}{\hspace{0mm}\shortstack{\Large KITTI \\ \Large rectified}}} &
\includegraphics[height=\turnheightnew, width=56mm]{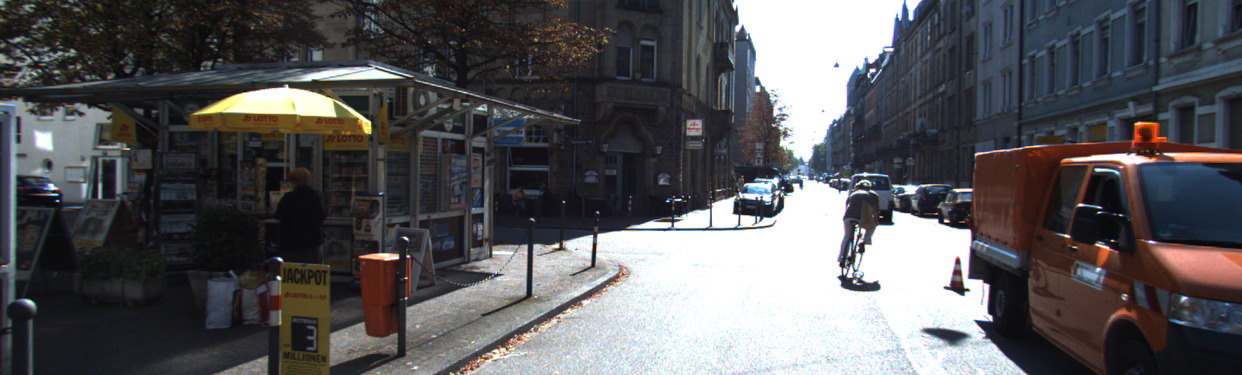} &
\includegraphics[height=\turnheightnew, width=56mm]{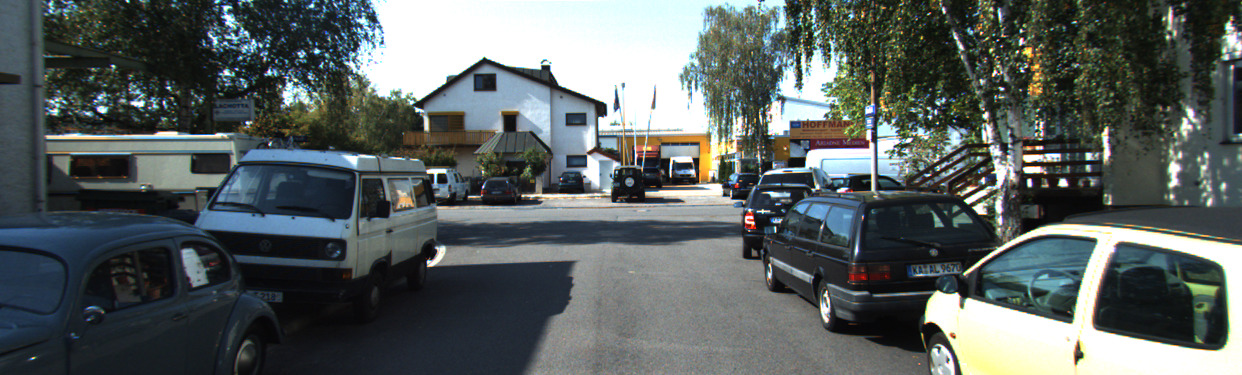} &
\includegraphics[height=\turnheightnew, width=56mm]{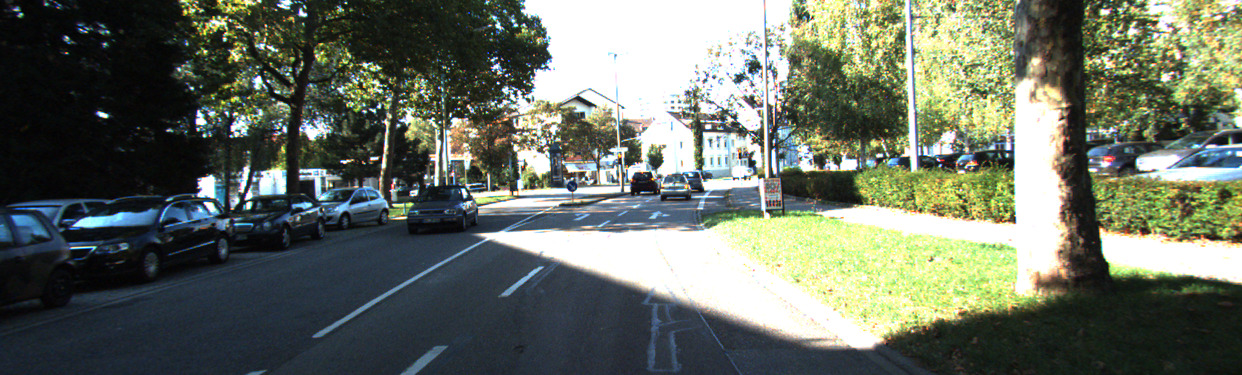} &
\includegraphics[height=\turnheightnew, width=56mm]{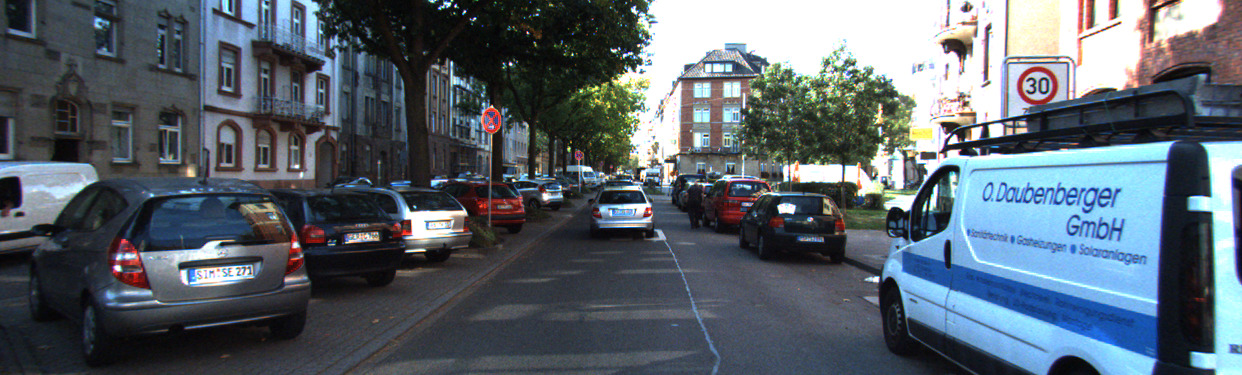}\\

{\rotatebox{90}{\hspace{0mm}\scriptsize}} &
\includegraphics[height=\turnheightnew, width=56mm]{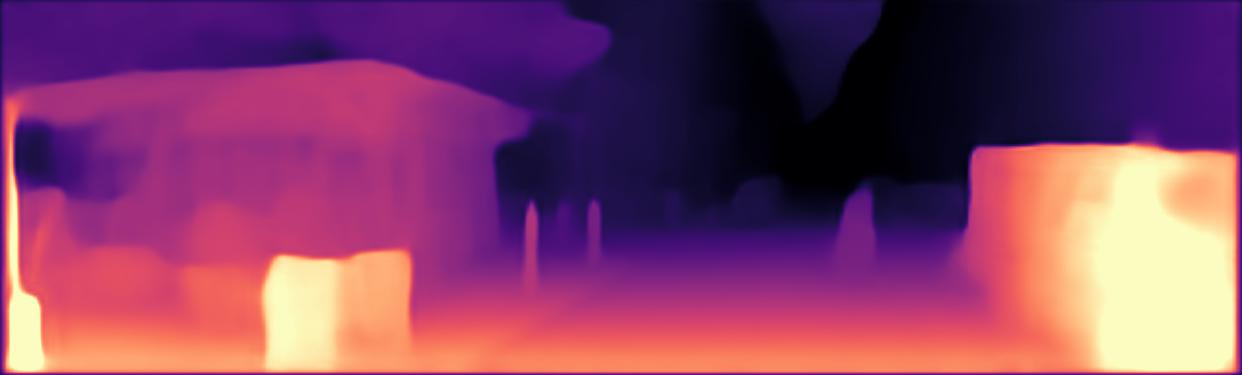} &
\includegraphics[height=\turnheightnew, width=56mm]{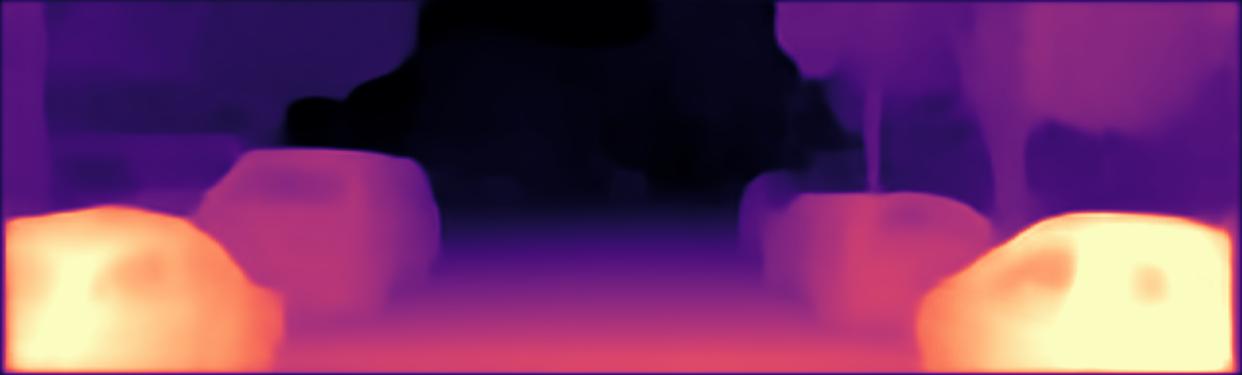} &
\includegraphics[height=\turnheightnew, width=56mm]{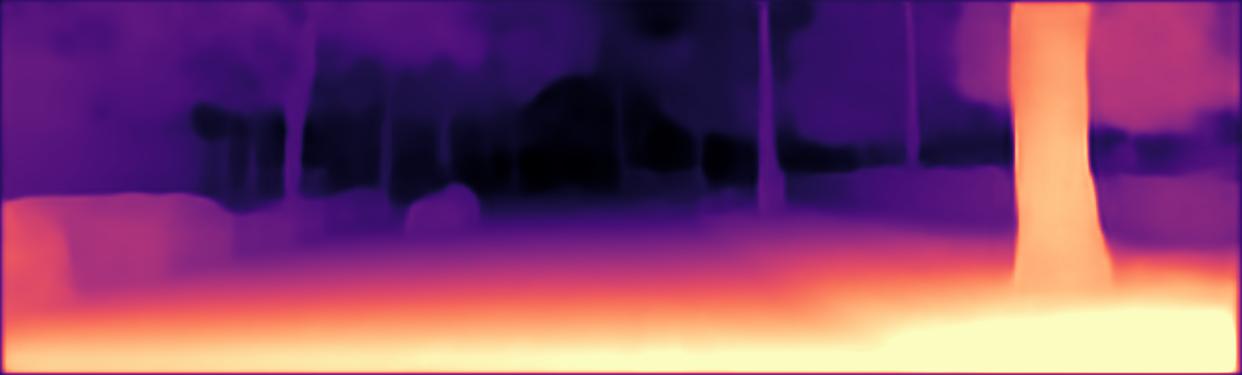} &
\includegraphics[height=\turnheightnew, width=56mm]{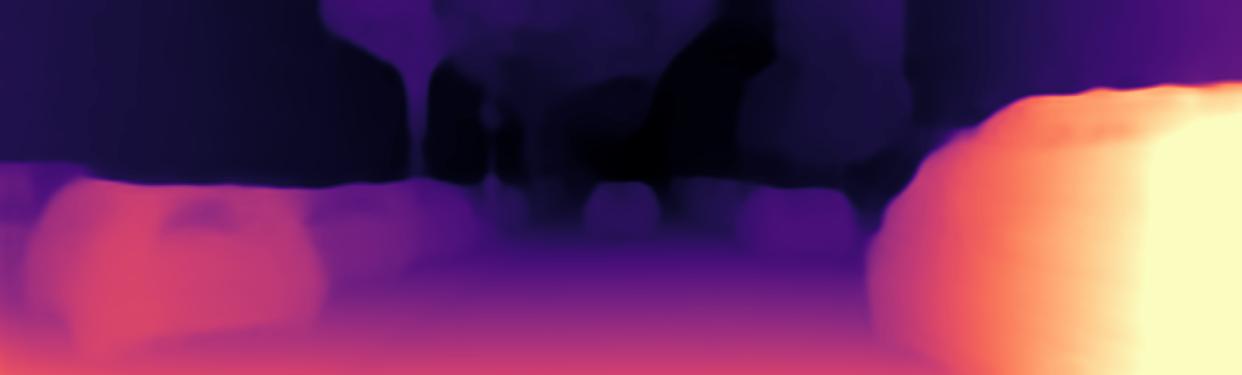} \\

{\rotatebox{90}{\hspace{0mm}\shortstack{\Large KITTI \\ \Large unrectified}}} &
\includegraphics[height=\turnheightnew]{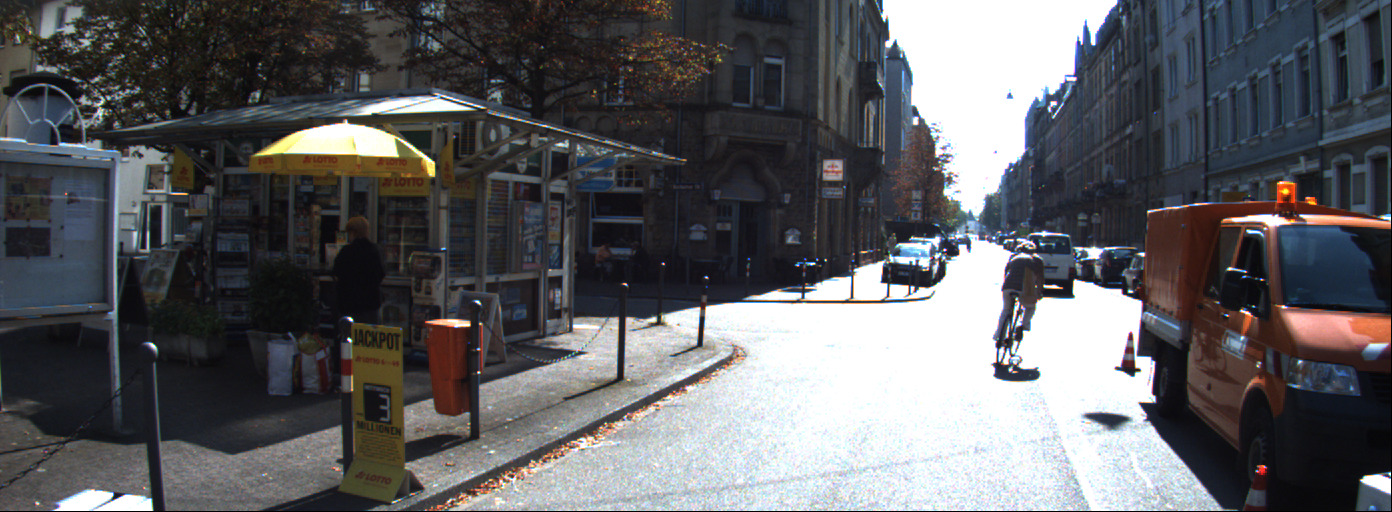} &
\includegraphics[height=\turnheightnew]{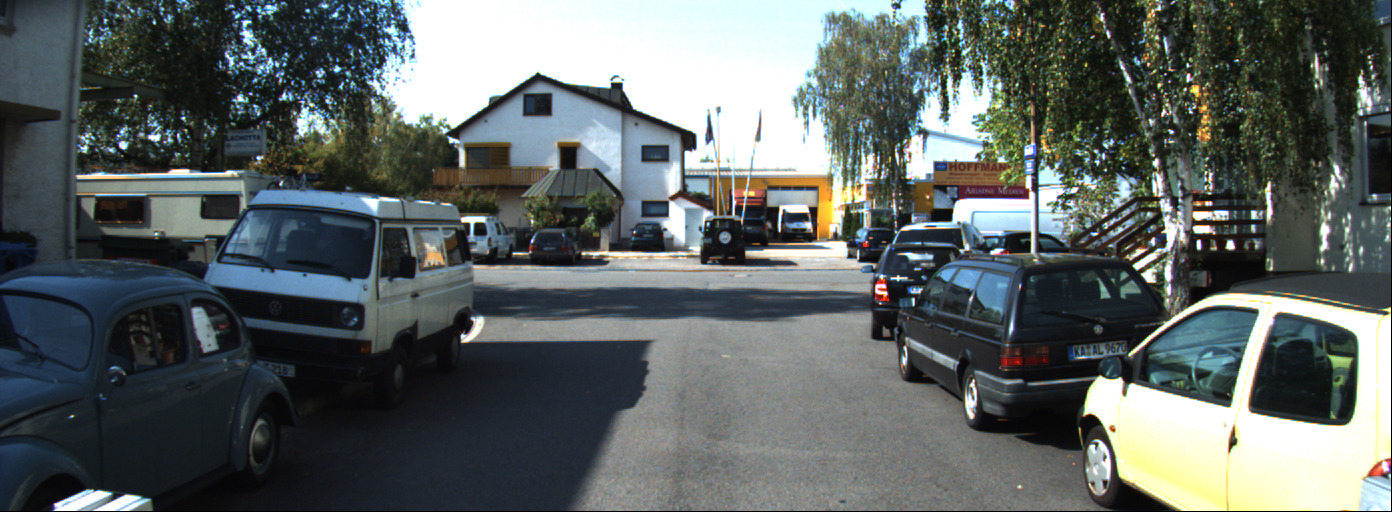} &
\includegraphics[height=\turnheightnew]{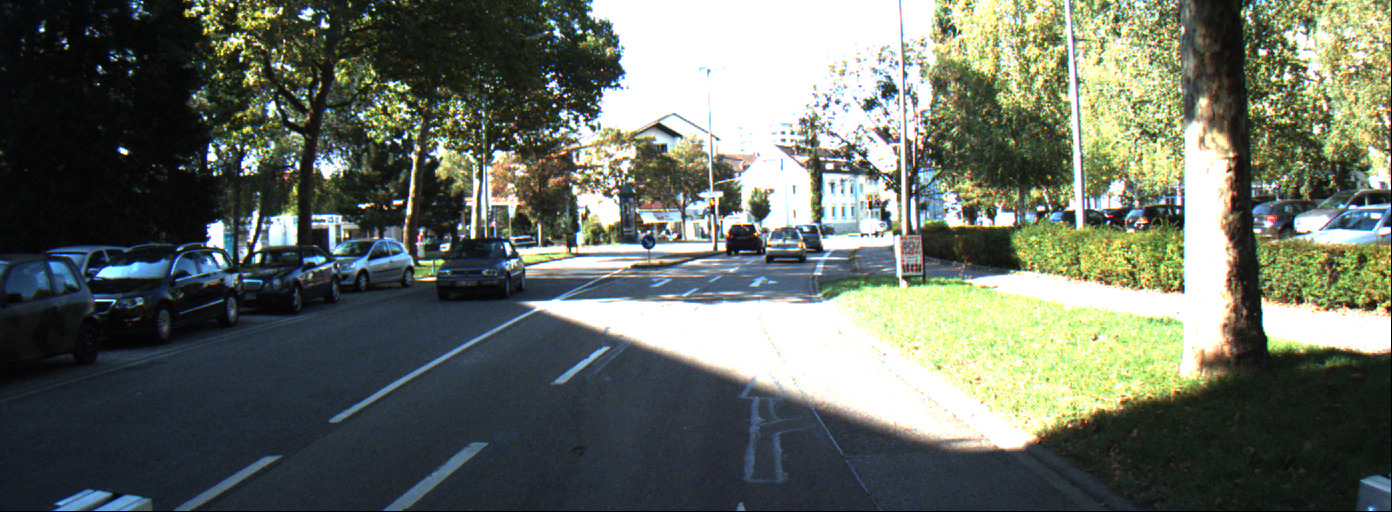} &
\includegraphics[height=\turnheightnew]{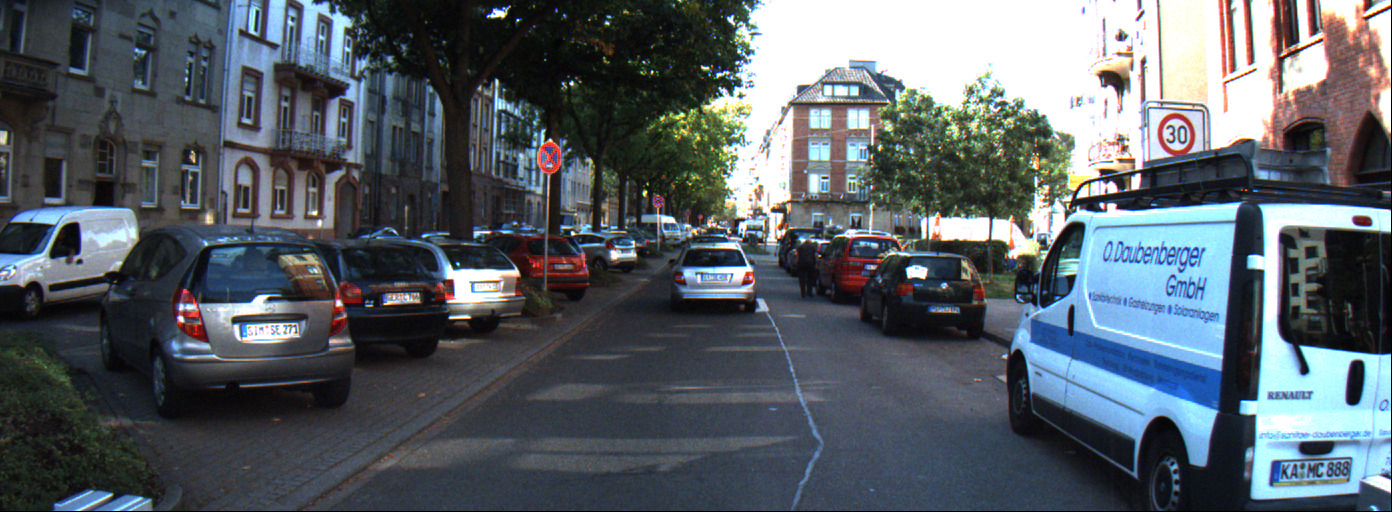}\\

{\rotatebox{90}{\hspace{0mm}\scriptsize}} &
\includegraphics[height=\turnheightnew]{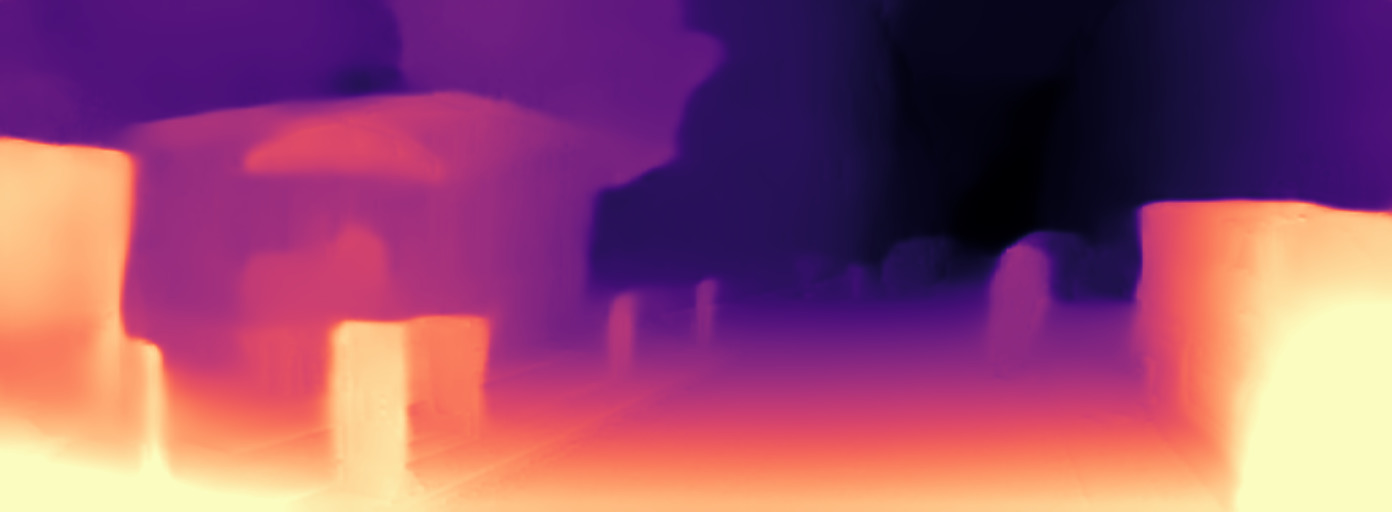} &
\includegraphics[height=\turnheightnew]{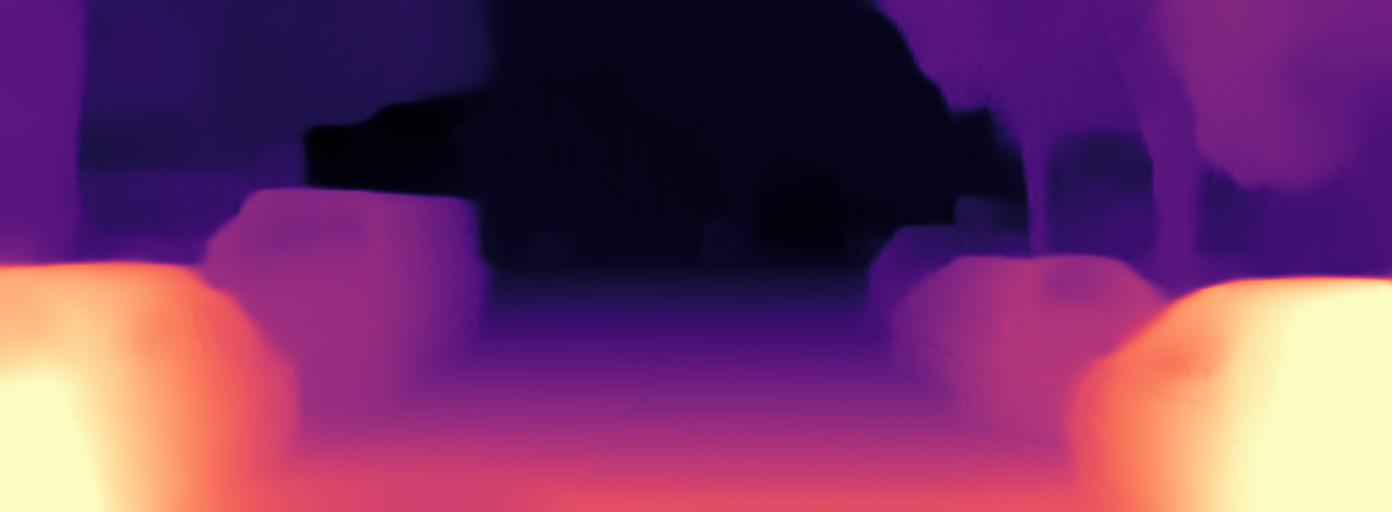} &
\includegraphics[height=\turnheightnew]{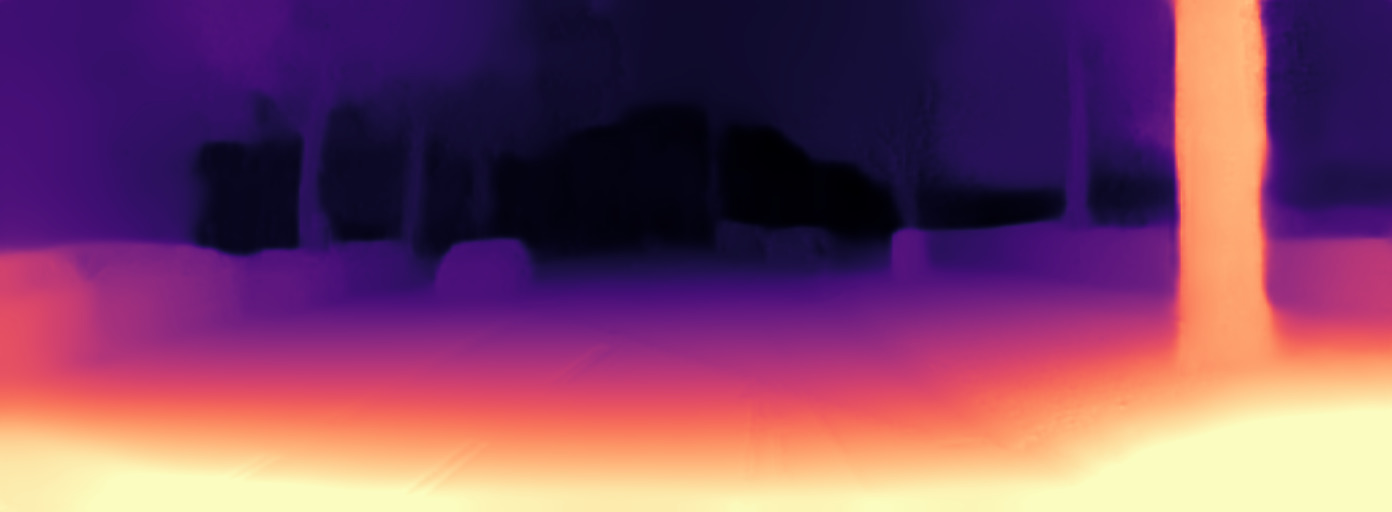} &
\includegraphics[height=\turnheightnew]{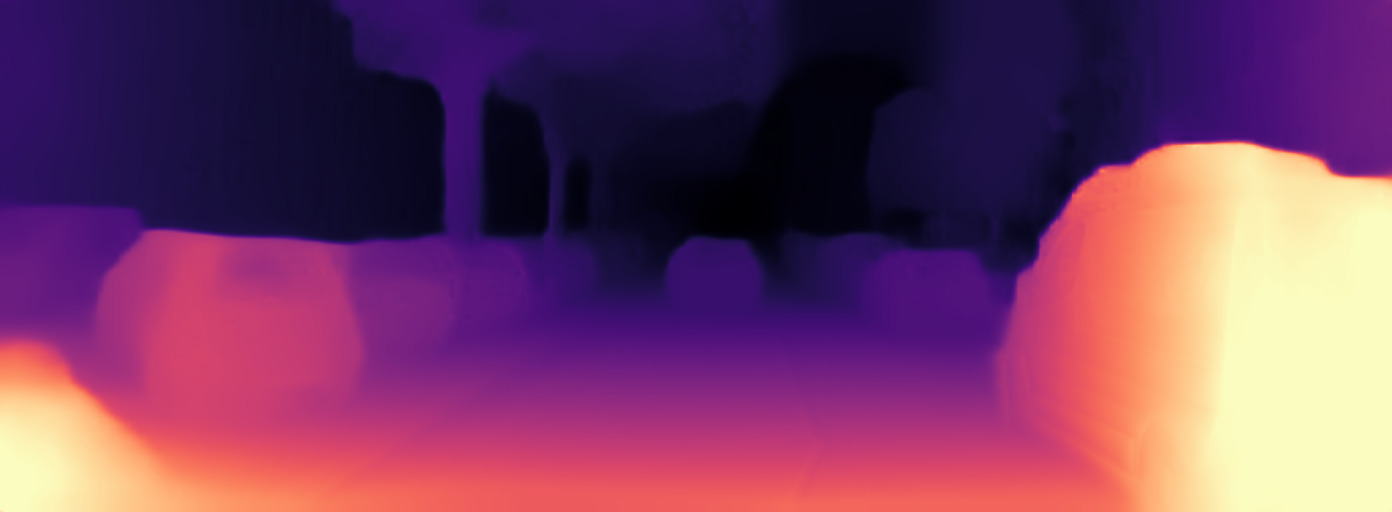} \\

{\rotatebox{90}{\hspace{0mm}\shortstack{\Large WoodScape \\ \Large cropped}}} &
\includegraphics[height=30mm, width=59mm]{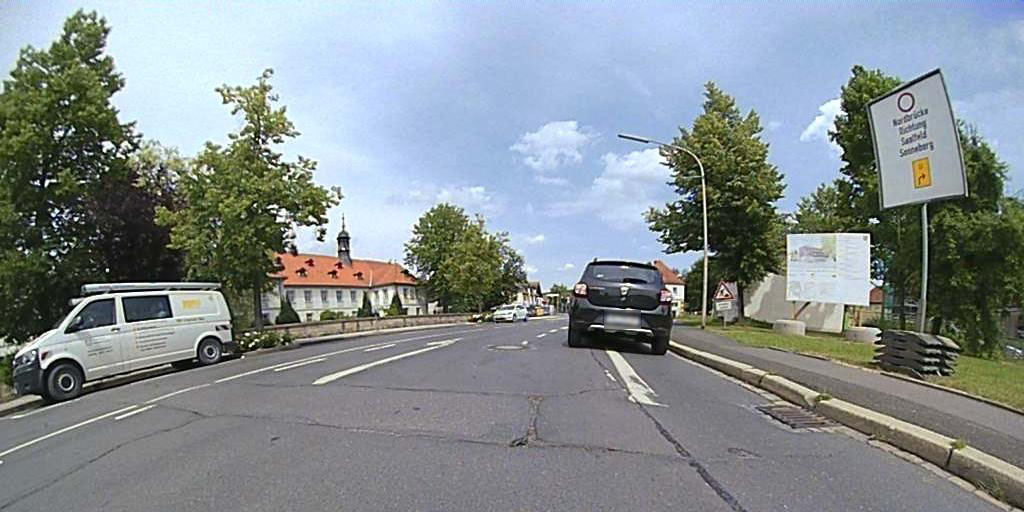} &
\includegraphics[height=30mm, width=59mm]{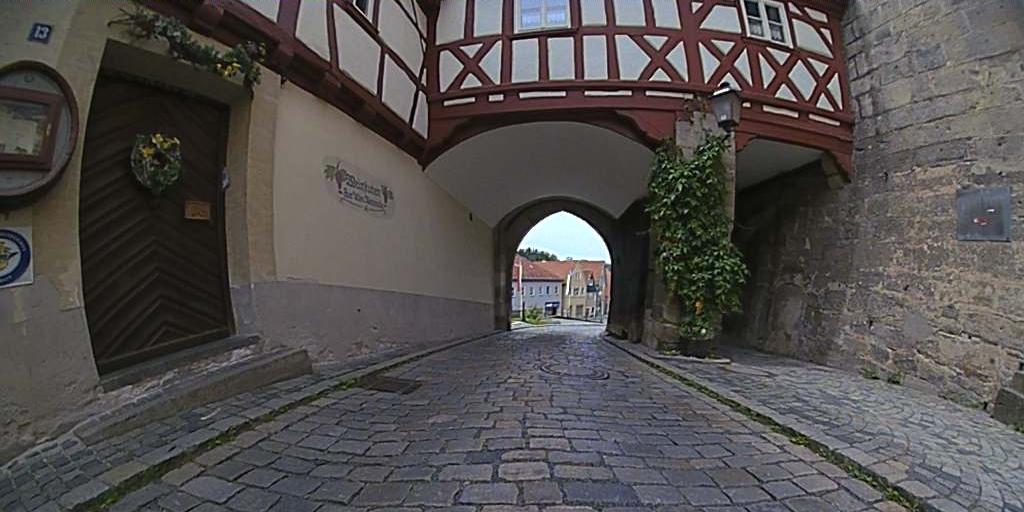} &
\includegraphics[height=30mm, width=59mm]{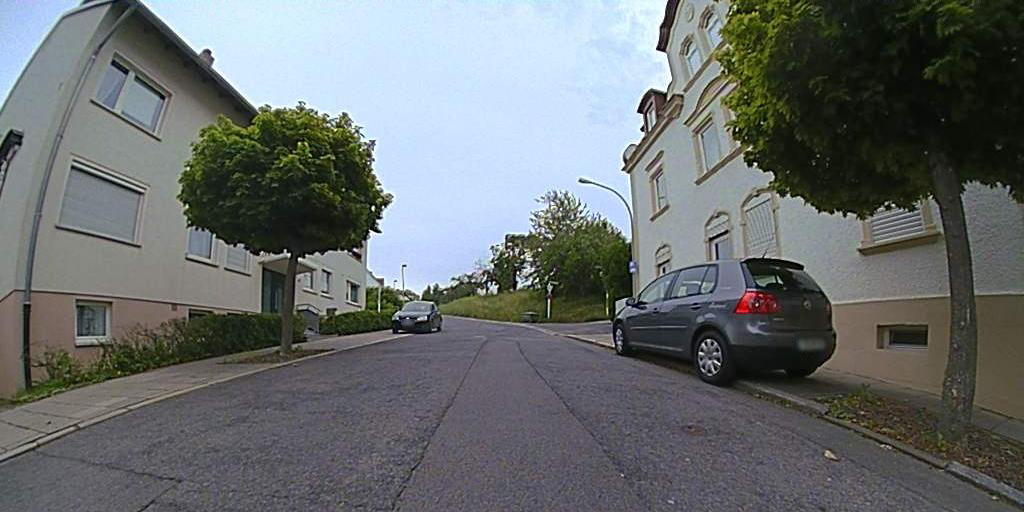} &
\includegraphics[height=30mm, width=59mm]{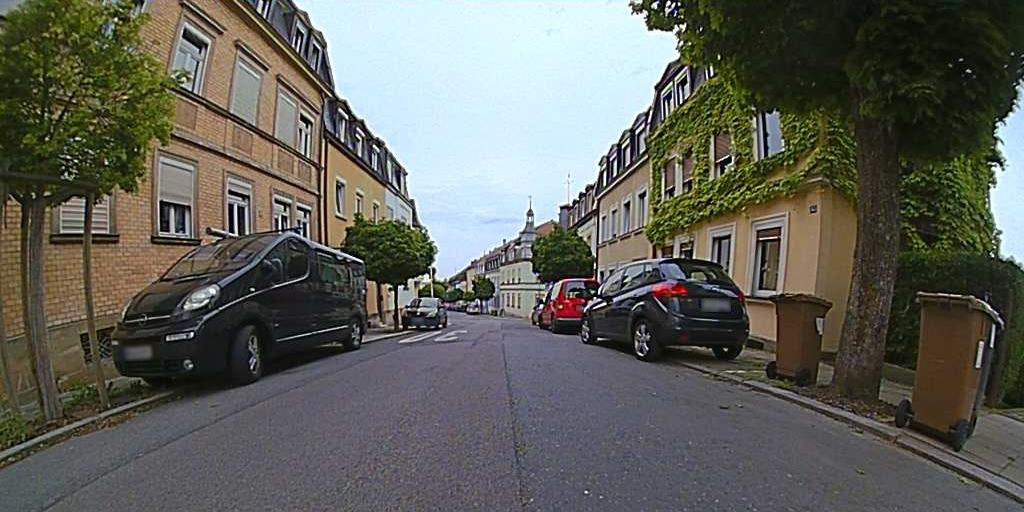}\\

{\rotatebox{90}{\hspace{0mm}}} &
\includegraphics[height=30mm, width=59mm]{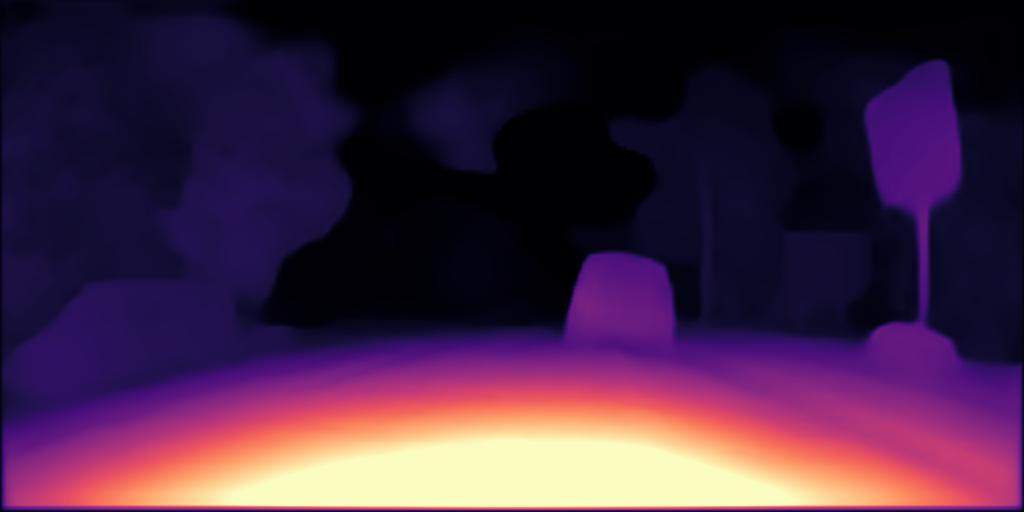} &
\includegraphics[height=30mm, width=59mm]{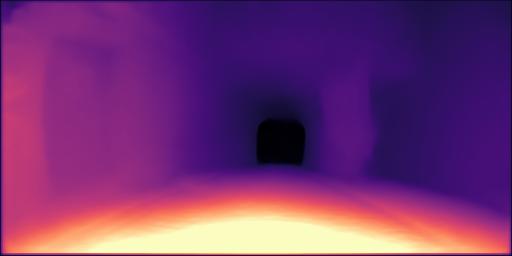} &
\includegraphics[height=30mm, width=59mm]{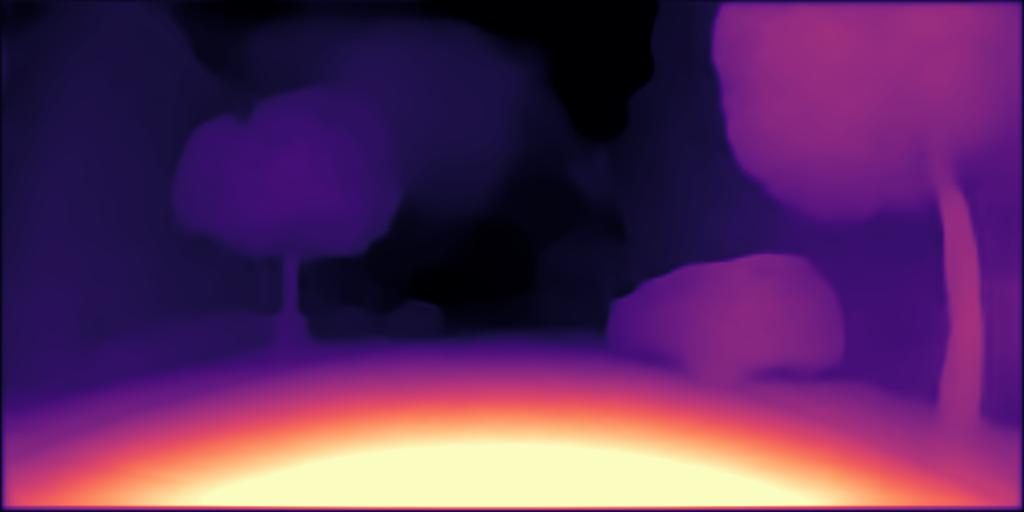} &
\includegraphics[height=30mm, width=59mm]{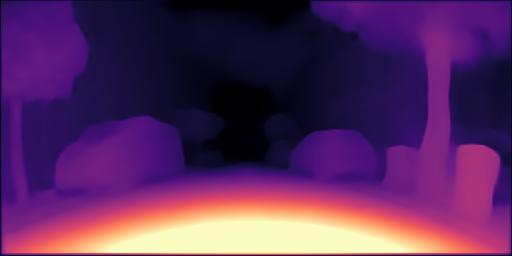}\\

\end{tabular}
}
  \caption{\textbf{Qualitative result comparison on KITTI and WoodScape dataset.} The results on a distorted test video sequence indicate excellent performance, see \url{https://youtu.be/K6pbx3bU4Ss}.}
  \label{fig:qual}
\vspace{-0.2cm}
\end{figure*}
\section{\textbf{Experiments}}

\textit{\textbf{Datasets -- KITTI and WoodScape}}
\label{sec:kitti_eigen_split}
For experiments with the Brown–Conrady model, the KITTI dataset is used as per the split defined by Eigen \etal~\cite{eigen2015predicting}. Following Zhou \etal~\cite{zhou2017unsupervised}, the static frames are dropped from the dataset. There are 39,810 images for training, 4,424 images for validation, and 697 images for testing. We also make use of the 652 test frames from the Eigen split with improved ground truth provided by~\cite{uhrig2017sparsity}. The WoodScape~\cite{yogamani2019woodscape} dataset distribution can be found in our FisheyeDistanceNet paper~\cite{kumar2019fisheyedistancenet}.\par
\textit{\textbf{Evaluation}}
To facilitate the comparison, we evaluate the results of UnRectDepthNet's depth estimation using the metrics proposed by Eigen \etal~\cite{Eigen_14}. Table~\ref{table:results} and Fig.~\ref{fig:qual} indicate the quantitative and qualitative results. The former illustrates that our scale-aware self-supervised approach on KITTI rectified outperforms almost all the state-of-the-art monocular approaches and the KITTI distorted results are better than most of the previous outcomes obtained with self-supervised approaches on the corresponding rectified dataset. Owing to the absence of odometry data, the Cityscapes dataset is not leveraged into our training framework.\par
Because the projection's operators are different, prior approaches to depth estimation will not work on the WoodScape fisheye dataset without significant redesign to incorporate fisheye projection geometry. In addition, fisheye cameras are designed for near-field sensing, and we only compare up to a range of 40\,m as per FisheyeDistanceNet~\cite{kumar2019fisheyedistancenet}. We generalized the training methodology of this model to incorporate any arbitrary distortion model. We also tuned our network to the optimal hyperparameters using grid search and removed batch normalization in the decoder as we observed ghosting effects and holes in homogeneous areas. We calculated the minimum reconstruction error for the two warps of the backward sequence individually compared to a combined minimization for forward sequence since here the target frames are $I_{t'}$ ($t^\prime \in \{t+1, t-1\}$), as explained in~\cite{kumar2019fisheyedistancenet}.\par 
\textit{\textbf{KITTI Distorted Ablation Study}}
We perform an ablation study to understand the significance of different components used and tabulate in Table~\ref{table:kittiablation}: (i) \textit{Remove Backward Sequence}: The network is trained only for a forward sequence consisting of two warps, as explained in~\cite{kumar2019fisheyedistancenet}. The impact is significant in the border areas as fewer constraints are induced. The model inherently fails to resolve unknown depths in those areas at the test time, which was also observed in previous works~\cite{godard2019digging,zhou2017unsupervised,yin2018geonet}; (ii) \textit{Additionally remove Super-Resolution using sub-pixel convolution}: It has a significant effect as distant objects are small in fisheye cameras and cannot be resolved correctly with simple nearest-neighbor interpolation or transposed convolution~\cite{odena2016deconvolution};
(iii) \textit{Additionally remove cross-sequence depth consistency loss}: The removal of the CSDCL diminishes the baseline, induces fewer constraints, and the model is not robust to yield accurate depth estimates.
\section{\textbf{Conclusion}}

We introduced a generic self-supervised training method for depth estimation handling distorted images. We support various commonly used automotive camera models in the framework and indicate empirical results on KITTI and WoodScape datasets. For KITTI, we show that depth estimation on unrectified images can produce the same accuracy as on rectified images. We also obtain a state-of-the-art result on KITTI among self-supervised algorithms. The same framework was used to train with fisheye unrectified images from the WoodScape dataset, where only the corresponding camera model parameters were updated. In future work, we aim to extend the training framework to take in various camera streams as input and output a generic inference model which can take in the camera model as argument.\\
\textbf{\textit{Acknowledgements}} 
We want to thank Valeo, especially DAR Kronach, Germany and Valeo Vision Systems, Ireland for supporting the creation of the WoodScape dataset. 
We want to thank Ciar\'an Eising (Valeo) and Ravi Kiran (Navya) for providing a detailed review.
\bibliographystyle{IEEEtran}
\bibliography{bib/ieee}
\clearpage
\end{document}